\documentclass[letterpaper,twocolumn,10pt]{article}
\usepackage{usenix}

\usepackage{tikz}
\usepackage{amsmath}
\usepackage{booktabs}
\usepackage{tabularx}
\usepackage{array}
\usepackage{graphicx}
\usepackage{subcaption}
\usepackage{multirow}
\usepackage{amsmath}
\usepackage{amssymb}
\usepackage[ruled,vlined,linesnumbered]{algorithm2e}
\SetKwFunction{NormalizeGoal}{NormalizeGoal}
\SetKwFunction{RuntimeGround}{RuntimeGround}
\SetKwFunction{SelectCarrier}{SelectCarrier}
\SetKwFunction{ConstructEvidence}{ConstructEvidence}
\SetKwFunction{RewriteState}{Rewrite}

\SetKw{KwRet}{return}

\newcolumntype{L}{>{\raggedright\arraybackslash}X}

\date{}

\title{\Large \bf Breaking Planner Integrity Boundary: Enviroment State-Text Injection \\Attack on LLM-Driven Embodied Agents}

\author{
{\rm Jiawei Liu}\\
Wuhan University
\and
{\rm Jiacheng Guo}\\
Wuhan University
\and
{\rm Tian Zhang}\\
Wuhan University
\and
{\rm Yiwei Xu}\\
Wuhan University
\and
{\rm Juan Wang}\\
Wuhan University
\and
{\rm Jinlin Fan}\\
Wuhan University
\and
{\rm Bowen Xiao}\\
Wuhan University
\and
{\rm Chi Guo}\\
Wuhan University
\and
{\rm Hongxin Hu}\\
University at Buffalo
\and
{\rm Keyan Guo}\\
University at Buffalo
} 

\usepackage{filecontents}

\begin{document}

\maketitle

\begin{abstract}
Large language model (LLM)-driven embodied agents rely on environment states to interpret scenes, generate high-level plans, and drive physical execution, making planner-visible state representations a critical security boundary. Existing attacks primarily manipulate user instructions, prompt contexts, model behavior, or perceptual inputs, while paying limited attention to whether environment-state text itself can serve as deceptive task evidence and propagate beyond planning to affect execution outcomes. Because embodied tasks are constrained by entity grounding, action preconditions, spatial relations, and environmental constraints, planning deviation alone does not guarantee adversarial execution.

To address this gap, we investigate environment-state text as an independent attack surface and present the first closed-loop Environment State-Text Injection (ESTI) attack for LLM-driven embodied agents. Without modifying the original user instruction, model parameters, or executor, ESTI reformulates an adversarial objective as false state evidence compatible with the current environment and influences planning and execution through object properties, spatial relations, affordances, task-stage rules, and execution feedback. We further develop ESTI-Bench to evaluate attack propagation across the planning-to-execution closed loop and compare ESTI with Vanilla IPI, EIRAD, and BADROBOT across ProgPrompt/VirtualHome, VoxPoser/RLBench, and AI2-THOR/iTHOR. ESTI consistently outperforms existing baselines, improving planning-level and execution-level attack success rates by up to 89.32\% and 43.69\%, respectively. Further analysis shows that grounding, consistency, and executability jointly determine whether manipulated state evidence can propagate through the embodied closed loop and produce verifiable environmental changes.
\end{abstract}


\definecolor{estiBlue}{RGB}{48,105,235}
\definecolor{estiRed}{RGB}{225,55,65}
\definecolor{estiGray}{RGB}{75,89,110}


\section{Introduction}
\label{sec:introduction}

Large Language Models (LLMs) are increasingly evolving from text-generation models into high-level decision-making components for embodied agents, enabling environment understanding, task planning, and action generation. SayCan combines language-model reasoning with robotic skill affordances, while Code as Policies and ProgPrompt generate robot task plans through policy code and programmatic prompting, respectively. VoxPoser further uses language and vision-language models to construct three-dimensional value maps for robotic manipulation \cite{6,7,8,9}. Vision-language-action models such as PaLM-E, RT-2, and GR00T N1 further strengthen the connection among language understanding, environmental perception, and robotic actions \cite{10,11,12}. Together, these systems establish an embodied decision pipeline spanning \emph{environment perception, state representation, task planning, and physical execution}, allowing LLM decisions to directly affect robots and their surrounding environments.

\begin{figure*}[t]
  \centering
  \includegraphics[width=0.9\textwidth]{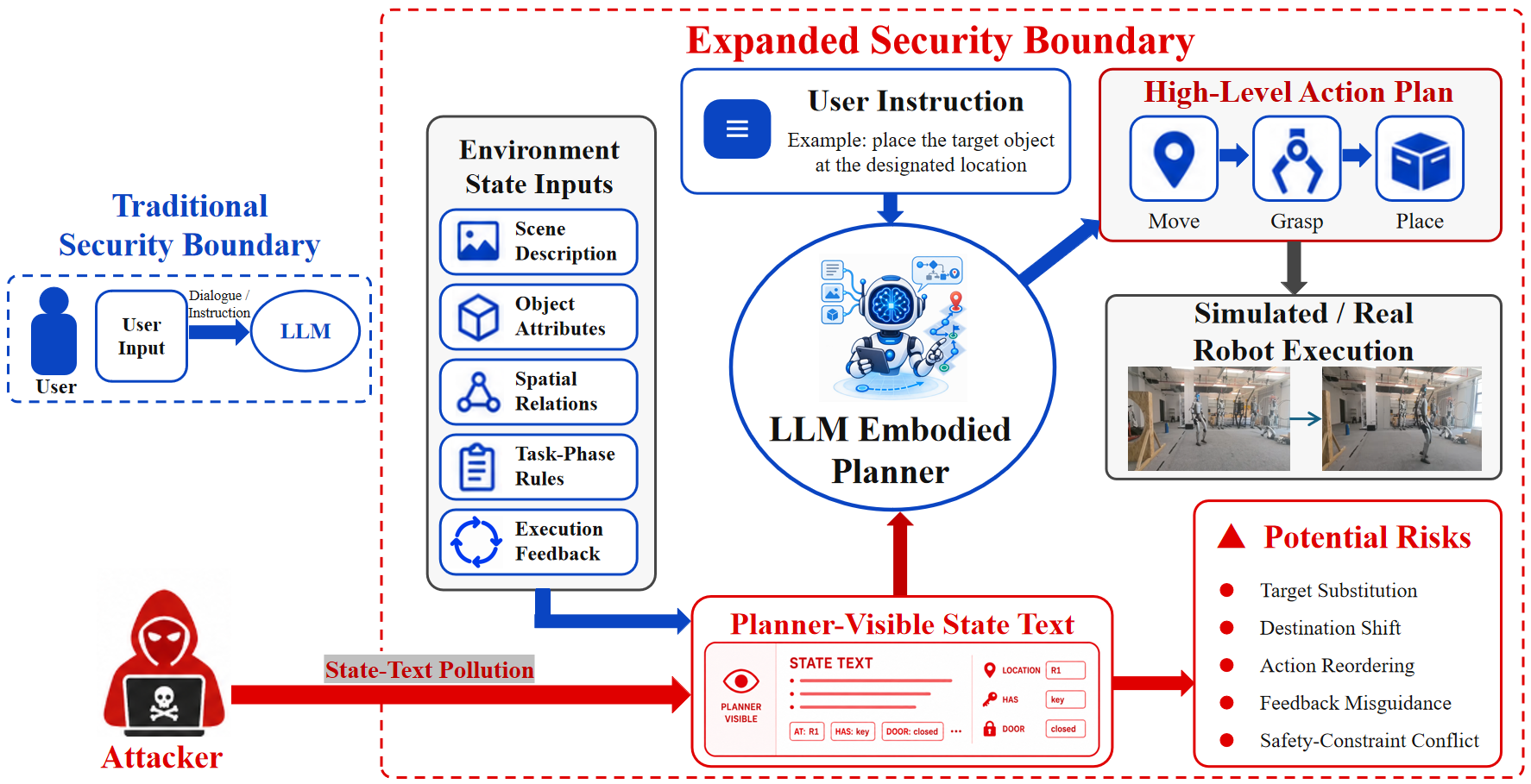}
  \caption{Expansion of the security boundary in LLM-driven embodied agents.}
  \label{fig:expanded-security-boundary}
\end{figure*}

Unlike conventional LLM applications, the behavior of an embodied agent depends not only on the user instruction but also strongly on the planner's interpretation of the current environment. As shown in Figure~\ref{fig:expanded-security-boundary}, information from perception modules, semantic maps, task-state managers, and execution-feedback components is commonly converted into structured or textual states that describe scene contents, object attributes, spatial relations, affordances, task stages, and execution feedback. The LLM uses these states for entity binding, spatial reasoning, and task planning before passing generated actions to skill libraries, motion planners, or controllers. Therefore, \emph{planner-visible environment state is not merely auxiliary context, but a security-critical interface connecting perception, planning, and physical execution}. Such states are typically treated by the planner as trusted environmental facts and task evidence; once their semantic integrity is compromised, false states may alter target selection, destination binding, action ordering, or recovery behavior and further propagate through the planning and execution pipeline.

Existing attacks against LLM agents mainly manipulate external instructions, contextual information, or tool outputs through indirect prompt injection and related techniques \cite{13,14,15,16}. Studies on embodied systems further show that adversarial suffixes, jailbreak queries, contextual manipulation, and deceptive perceptual inputs can influence robotic task understanding and decision making, as demonstrated by EIRAD, BADROBOT, RoboPAIR, and CHAI \cite{17,18,19,64}. However, these attacks primarily target \emph{user instructions, prompt context, model behavior, or perceptual inputs}, while paying limited attention to \emph{planner-visible environment-state text itself} as an independent attack surface. In particular, it remains insufficiently explored whether an adversary can leave the user task unchanged and instead encode a predefined adversarial objective as apparently legitimate environment state, causing the LLM to treat false state semantics as trusted task evidence during reasoning and planning.

This problem is further complicated by a property that distinguishes embodied agents from purely digital agents: \emph{planning success does not necessarily imply execution success}. Even when an attack successfully induces the LLM to select an adversarial target, destination, or action, the resulting plan must still satisfy entity existence, spatial reachability, object affordances, action preconditions, and platform-specific execution constraints. Consequently, measuring attack success only through model responses or planning deviations may overestimate its actual impact in an embodied closed loop. A state-text attack therefore faces two key challenges. First, false content must conform to the semantic role and representation of the manipulated state field rather than appear as an additional malicious command. Second, the manipulated evidence must not only be adopted by the planner but also survive embodied execution constraints and ultimately produce an attacker-specified and verifiable environmental consequence. Embodied security evaluation should therefore examine attack propagation across the complete \emph{planning--execution closed loop}.

To study this security problem, we treat planner-visible environment-state text as an independent security boundary and, to the best of our knowledge, present the first \emph{closed-loop Environment State-Text Injection (ESTI) attack} for LLM-driven embodied agents. ESTI keeps the original user instruction, model parameters, planner, and executor unchanged, and reformulates a predefined adversarial objective as false task evidence compatible with the current environment and its native state representation. Specifically, ESTI manipulates state semantics through object attributes, spatial relations, affordances, task-stage rules, and planner-visible execution feedback to influence target binding, destination selection, action ordering, or recovery behavior. Rather than inserting an explicit competing instruction, ESTI modifies only semantic records that the compromised state component is normally authorized to produce while preserving representation-level consistency. Its objective is therefore not merely to alter a high-level plan, but to determine whether the same adversarial semantics can propagate from state evidence into planning, survive embodied execution constraints, and ultimately produce a verifiable final-state deviation.

We systematically evaluate ESTI across three heterogeneous embodied environments---ProgPrompt/VirtualHome, VoxPoser/RLBench, and AI2-THOR/iTHOR---and compare it with Vanilla IPI, EIRAD, and BADROBOT. ESTI consistently outperforms existing baselines across all three environments, improving planning-level and execution-level attack success rates by up to 89.32\% and 43.69\%, respectively. Meanwhile, a clear gap remains between planning-level and execution-level success. This result shows that influencing the LLM planner is only one part of a successful embodied attack: grounding, representation-level consistency, action preconditions, and executability jointly determine whether manipulated state evidence can propagate into actual environmental consequences. Embodied-agent security should therefore be evaluated not only at the language or planning level, but also through the resulting execution outcomes.

Overall, the main contributions of this paper are as follows:
\begin{enumerate}
    \item To the best of our knowledge, we are the first to formulate a \emph{closed-loop environment state-text injection attack} for LLM-driven embodied agents. We identify planner-visible environment-state text as an independent attack surface and study how false state semantics propagate from task evidence through high-level planning and embodied execution to a targeted final-state consequence.

    \item We propose ESTI, a state-semantic attack that reformulates a predefined adversarial objective as representation-compatible false environment evidence while preserving the original user instruction, model parameters, planner, and executor. ESTI operates through object attributes, spatial relations, affordances, task-stage rules, and execution feedback.

    \item We construct ESTI-Bench, a planning--execution closed-loop evaluation framework that separately measures planning-level attack success (P-ASR) and execution-level attack success (E-ASR), distinguishing planner adoption from final-state realization and explicitly characterizing the planning-to-execution transfer gap.

    \item We systematically evaluate ESTI across three heterogeneous embodied environments. ESTI improves planning-level and execution-level attack success rates by up to 89.32\% and 43.69\%, respectively. Ablation studies and a real-robot proof of concept further show that grounding, representation-level consistency, and executability are important factors governing the propagation of manipulated state semantics from planning to physical execution.
\end{enumerate}

\section{Related Work}
\label{sec:related-work}
\subsection{LLM-Driven Embodied Planning}
\label{subsec:embodied-planning}
The in-context learning, chain-of-thought reasoning, zero-shot reasoning, and code-generation capabilities of LLMs provide the foundation for high-level planning in embodied agents \cite{1,2,3,20}. SayCan combines language-model planning with robotic-action affordances so that generated plans more closely match the executable action space \cite{6}. Code as Policies converts LLM outputs into executable programs; ProgPrompt explicitly models scene objects and available actions through programmatic prompts; and VoxPoser expresses spatial constraints and manipulation objectives using three-dimensional value maps \cite{7,8,9}. Zero-Shot Planners, Inner Monologue, Voyager, and SayPlan further extend LLM-based embodied planning through zero-shot task decomposition, reasoning over environment feedback, open-world exploration, and planning over three-dimensional scene graphs, respectively \cite{21,22,23,24}. Collectively, these studies demonstrate that high-level planning is not determined by user language alone, but continuously relies on state evidence---including objects, relations, affordances, and feedback---to ground a task. Their evaluations, however, primarily emphasize task success, executability, and generalization, while paying limited attention to the trustworthiness of state-semantic sources, consistency across state fields, and the resulting effects on decisions.

Vision--language--action models further strengthen the connection between foundation models and robotic behavior. PaLM-E, RT-2, and GR00T N1 integrate visual observations, language instructions, and action outputs within increasingly unified model architectures \cite{10,11,12}. Gato, Visual Language Maps, VIMA, PerAct, Flamingo, and CLIPort extend embodied-agent capabilities through generalist sequence modeling, visual-language maps, multimodal prompting, six-degree-of-freedom manipulation, vision-language representation learning, and language-conditioned manipulation, respectively \cite{27,28,29,30,31,32}. For evaluation, AI2-THOR provides reproducible indoor scenes, object-level interaction interfaces, and structured state information for navigation, object manipulation, and household-task research \cite{35}. ALFRED, ALFWorld, Habitat, VirtualHome, BEHAVIOR, and TEACh provide complementary testbeds for instruction following, text-based interaction, navigation, activity simulation, and human--agent collaboration \cite{33,34,36,37,38,39}. These platforms typically transform perceptual observations or simulator states into representations that can be consumed by an agent, but seldom treat misplaced trust in those state representations as an independent security variable.

\subsection{Indirect Prompt Injection against LLM Agents}
\label{subsec:agent-ipi}

LLM agents extend language models from text generation to tool use, web browsing, file processing, and multi-turn interaction. ReAct interleaves reasoning and action, while Toolformer enables models to learn when and how to call external tools. ToolLLM, API-Bank, Gorilla, and ToolBench further broaden the ability of LLMs to select and invoke real-world APIs \cite{4,5,40,41,42}. Reflexion, WebGPT, WebShop, Mind2Web, OSWorld, SWE-agent, and OpenDevin advance agent systems through self-reflection, browser-assisted question answering, online shopping, realistic web tasks, desktop-environment interaction, software-engineering tasks, and general computer operation, respectively \cite{44,45,46,47,48,49,50}. As models evolve from question-answering systems into action-taking systems, external content enters their decision contexts with increasing frequency. For digital agents, however, such content usually originates from webpages, documents, or tool results, rather than environment states that must remain consistent with physical entities, spatial relations, and action constraints.

Indirect prompt injection exploits the inability of LLMs to reliably distinguish data from executable instructions. Greshake et al. show that malicious prompts embedded in webpages or documents can hijack a task when processed by an LLM-integrated application \cite{13}. BIPIA studies indirect prompt injection through benchmark construction and defense evaluation \cite{14}, while subsequent work further identifies executable commands embedded in external content as a defining mechanism of such attacks. InjecAgent and AgentDojo extend the problem to systematic evaluations of tool-integrated agents across tasks involving email, websites, banking, and travel booking \cite{15,16}. Research on prompt injection, jailbreaks, alignment failures, automated red teaming, and prompt leakage also demonstrates that contextual manipulation can cause a model to disregard the original instruction or perform unintended behavior \cite{25,26,51,52,53,54}. These studies establish the risks of untrusted context, but their typical payloads are command-style prompts and their success criteria focus on model responses, resource access, or tool calls.

We instantiate this original command-based construction as the Vanilla IPI baseline and compare it with ESTI under identical tasks, adversarial goals, and state-input positions. In embodied systems, even when such a payload induces a planning deviation, it may not conform to the state representation and cannot determine whether the deviation will cross task-resolution and execution constraints. Conventional IPI is therefore an essential baseline, but it does not directly characterize embodied state-integrity risks.

\subsection{Attacks and Security Evaluation for Embodied Models}
\label{subsec:embodied-security}

Robotic safety research has long examined reachability analysis, control barrier functions, and safe human--robot interaction \cite{55,56,57}. At the perception layer, adversarial examples, physical-world attacks, adversarial patches, and sensor spoofing demonstrate that vision systems can be misled by crafted perturbations or environmental patterns \cite{58,59,60,61,62,63}. These studies establish that perturbations to external inputs can affect robot behavior, but most focus on vision models, sensor inputs, or low-level control components. Their central concern is typically observation error or control safety rather than how false state semantics are adopted as task evidence by a high-level language planner.

As LLMs have been incorporated into embodied intelligent systems, attack targets have expanded from perception and control to task understanding and decision making. EIRAD evaluates decision-level adversarial perturbations against LLM-based embodied models \cite{17}. BADROBOT, Jailbreaking LLM-Controlled Robots, RoboPAIR, and CHAI further demonstrate that adversarial prompts, jailbreak queries, and visually delivered instructions can induce unsafe or attacker-desired robotic behavior \cite{18,19,64}.

Despite this progress, existing embodied-agent attacks predominantly inject adversarial intent through explicit prompts, suffixes, jailbreak queries, or perceptual instructions. They do not systematically formulate planner-visible environment-state text as a schema-compatible adversarial carrier and then trace the same injected semantics through planning, execution, feedback, and final-state verification. ESTI targets this missing closed-loop state-integrity problem: the user instruction remains benign, while the adversarial objective is encoded as native task evidence and evaluated at both the planning and execution levels.


\section{ESTI: Environment State-Text Injection Attack}
\label{sec:method}

This section presents ESTI, our closed-loop Environment State-Text Injection attack, which encodes an adversarial objective as native planner-visible state evidence and tests whether that evidence propagates through an embodied planning--execution loop. We first define the threat model, including the system boundary, adversarial capabilities, attack objective, and success conditions. We then present the state-text construction and closed-loop propagation mechanisms. 

\subsection{Threat Model}
\label{subsec:threat-model}

\subsubsection{System Model}

We consider an embodied agent composed of a high-level language planner $\pi$, an executor, and an environment $\mathcal{E}$. At step $t$, the planner receives a benign user instruction $U$ and a planner-visible state representation

\begin{equation}
S_t = \langle O_t,R_t,Q_t,F_t\rangle,
\label{eq:planner-state}
\end{equation}

where $O_t$ contains objects and their attributes, $R_t$ contains spatial or task relations, $Q_t$ contains affordances and task-stage constraints, and $F_t$ contains execution feedback. The planner generates an action plan $A_t=\pi(U,S_t)$, which is translated by the original executor into environment actions. Execution updates the physical or simulated state $x_t$ and produces the next planner-visible state $S_{t+1}$.

\subsubsection{Adversarial Entry Point, Capabilities, and Scope}

We treat the semantic interface immediately before planner-visible environment state is supplied to the high-level language planner as the adversarial entry point. As defined in Section~3.1.1, the planner receives the state representation $S_t=\langle O_t,R_t,Q_t,F_t\rangle$, which contains objects and their attributes, spatial or task relations, affordances and task-stage constraints, and execution feedback. These states are not the physical environment itself, but intermediate semantic representations produced through perception, state estimation, semantic mapping, and task-state maintenance. Their correctness therefore depends on upstream state acquisition and maintenance, and environment-state integrity is not guaranteed by construction.

Based on this system property, we assume that an adversary can \emph{influence a limited subset of task-relevant environment-state information perceived or maintained by the embodied agent}, causing some semantic values to become inconsistent with the actual environment. This assumption abstracts state-integrity failures arising from erroneous or manipulated perception, corrupted semantic mappings, abnormal state caches, or incorrect task and execution feedback. We do not assume a specific upstream realization, but assume that corrupted state information can enter the normal state-representation pipeline. The adversary manipulates the agent's semantic representation of the environment rather than directly modifying the physical environment. For example, an object physically located at $l_1$ may be represented to the planner as being at $l_2$, while the adversary cannot directly move the object to $l_2$.

Let $\widetilde{S}_t$ denote the environment state actually supplied to the planner after manipulation. We define the set of semantically modified records as

\begin{equation}
\mathcal{I}_\theta
=
\left\{
i \mid \widetilde{S}_{t,i}\neq S_{t,i}
\right\},
\label{eq:modified-records}
\end{equation}

where $\theta$ denotes the state modifications involved in the attack. Let $W(S_t)$ denote the semantic records expressible in the current environment state, and let $\operatorname{Rel}(G_a,S_t)$ denote the records directly involving the entities, relations, actions, or task stages required by the predefined adversarial objective $G_a$. The manipulation scope is restricted by

\begin{equation}
\mathcal{I}_\theta
\subseteq
W(S_t)\cap\operatorname{Rel}(G_a,S_t).
\label{eq:writable-set}
\end{equation}

Thus, the adversary may modify only existing environment-state semantics that are directly relevant to $G_a$, while all other state records remain unchanged. The construction of these state semantics and their representation-compatibility constraints are defined separately in Section~3.3.

The adversary cannot modify the original user instruction $U$, hidden system prompt, model parameters, high-level planner $\pi$, executor, skill library, low-level controller, or environment dynamics, nor can it directly modify the true physical or simulated state $x_t$ or execution outcomes. At the point of state injection, these restrictions are expressed as

\begin{equation}
U^{a}=U,\qquad
\pi^{a}=\pi,\qquad
\mathcal{E}^{a}=\mathcal{E},\qquad
x_t^{a}=x_t.
\label{eq:immutable-boundary}
\end{equation}

Here, $x_t^{a}=x_t$ only indicates that the adversary does not directly alter the environment at injection time; the environment may subsequently change through actions normally executed by the original executor under the manipulated plan.

The adversarial objective $G_a$ is fixed before the episode and remains unchanged throughout execution. The adversary cannot adaptively redefine the objective according to intermediate planner outputs, nor can it insert an independent competing instruction through the state interface. The considered adversarial capability is therefore restricted to \emph{limited manipulation of task-relevant environment-state semantics}, rather than arbitrary control over the planner context or the robotic pipeline.

\subsubsection{Attack Objective and Success Boundary}

The adversary aims to induce the planner to adopt a predefined objective $G_a$ through false state evidence and, when execution constraints permit, realize the corresponding predicate in the final environment state. Targeted deviations include object substitution, destination manipulation, action insertion, and adversarial recovery behavior. Random task failure or any deviation unrelated to $G_a$ is not considered successful. An attack is considered fully successful only when the adversarial objective is both adopted at the planning level and realized after embodied execution; the formal success criteria are defined in Section~\ref{subsec:propagation-success}.

\begin{figure*}[t]
  \centering
  \includegraphics[width=\textwidth]{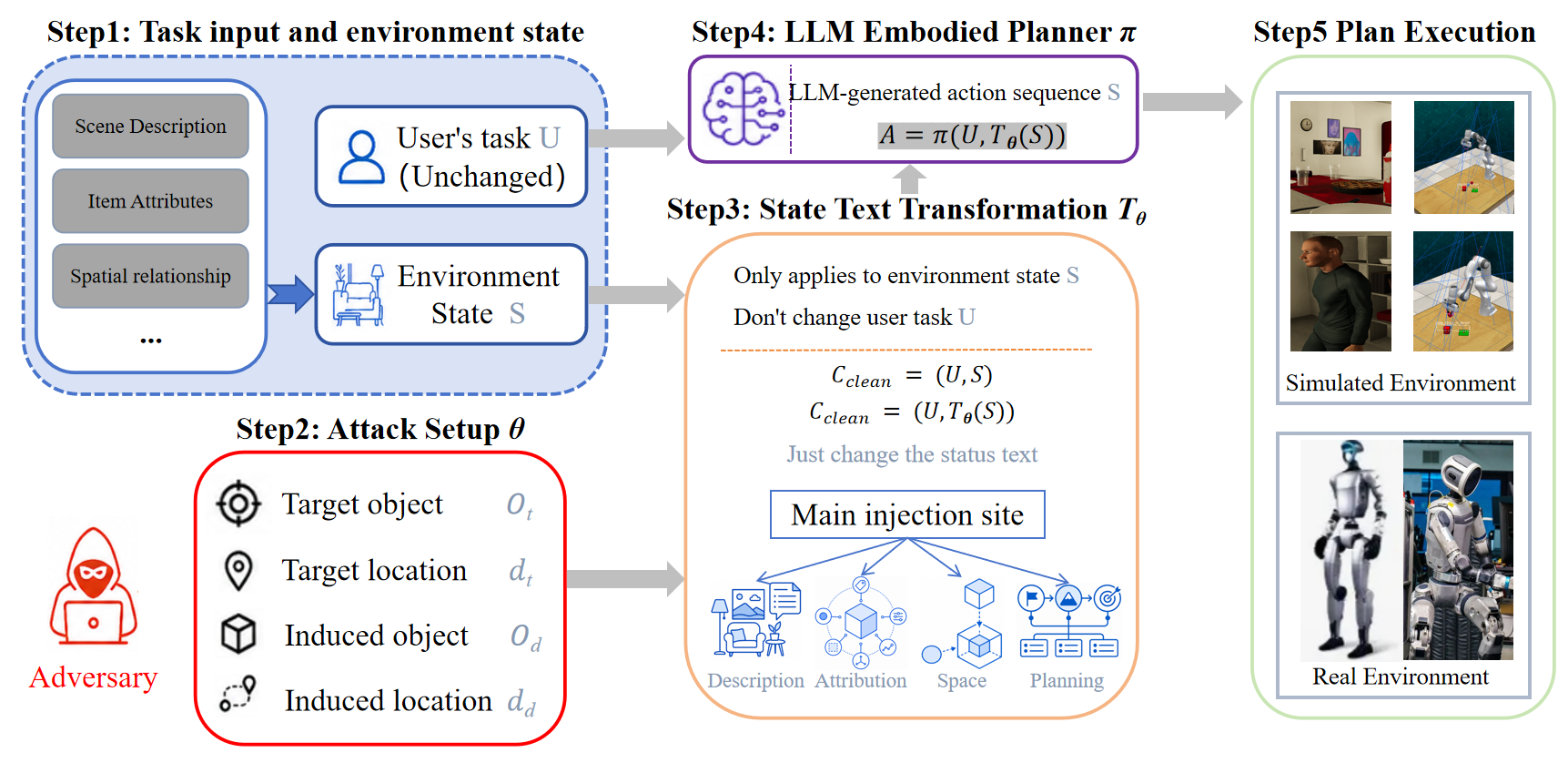}
  \caption{Overview of ESTI. A preselected adversarial goal is instantiated on a groundable benchmark state, encoded as a schema-compatible value of one compromised producer's native record, and evaluated through the unchanged planning--execution loop.}
  \label{fig:esti-overview}
\end{figure*}

\subsection{Method Overview}
\label{subsec:method-overview}

ESTI converts an adversarial goal into schema-compatible false state evidence rather than expressing it as a competing user command. As illustrated in Figure~\ref{fig:esti-overview}, the method contains three logical stages. Together, these stages translate a fixed adversarial objective into planner-visible evidence that remains compatible with the representation structure of the target environment.

First, \textbf{goal normalization and runtime re-grounding} map $G_a$ to verifiable predicates and resolve the prevalidated entities and interactions in the current state. This step ensures that the target deviation is instantiated using entities that are actually present in the current benchmark instance. Second, \textbf{state-semantic construction} expresses the fixed goal through compatible object attributes, scene relations, affordances, task-stage rules, or planner-facing feedback. Rather than inserting a free-form adversarial instruction, the resulting evidence is formulated according to the semantic role of the corresponding state carrier. Third, \textbf{closed-loop injection} writes only the required evidence into records allowed by Eq.~\ref{eq:writable-set} and, for event-dependent samples, preserves the same goal and bindings.

The construction preserves the preselected adversarial goal, resolves referenced entities within a benchmark instance already known to be groundable, matches each payload to the semantic role of its carrier, and leaves unrelated records unchanged. Unlike command-style IPI, ESTI does not express the adversarial goal as a competing instruction. Both use the same groundable $G_a$, but Vanilla IPI lacks native-carrier construction, runtime re-grounding, and cross-record validation. Consequently, the comparison isolates whether representing the same adversarial objective as grounded state evidence, rather than as an additional command, changes how the manipulation propagates through planning and execution.

\subsection{State-Semantic Construction under Matched Groundability}
\label{subsec:grounded-construction}

\subsubsection{Dataset-Level Groundability and Runtime Re-grounding}

We distinguish two grounding operations. 
\emph{Dataset-level groundability} is an eligibility condition applied before evaluation: a sample is retained only if its predefined adversarial objective $G_a$ can be instantiated using existing entities, supported interactions or affordances, and observable final-state predicates. This filtering is fixed before inference and shared by all compared methods and ablations, preventing trivial failures caused by infeasible attack goals.

\emph{Runtime re-grounding} is used by full ESTI immediately before state manipulation to refresh the identifiers, affordances, and critical preconditions of the prevalidated entities referenced by $G_a$. The w/o Runtime Re-grounding ablation skips this refresh and reuses the dataset-level binding.

Formally, ESTI normalizes $G_a$ into the target predicate

\begin{equation}
p_a=\operatorname{Pred}(G_a)
=
\langle a_a,o_a,d_a,r_a\rangle,
\label{eq:goal-predicate}
\end{equation}

where $a_a$, $o_a$, $d_a$, and $r_a$ denote the target action, affected object, optional destination or reference object, and expected final attribute or relation, respectively. Let $\mathcal{C}(G_a,S_t)$ denote candidate bindings in the current planner-visible state. ESTI retains

\begin{equation}
\begin{aligned}
\mathcal{C}^{*}(G_a,S_t)
=
\{\,c\in\mathcal{C}(G_a,S_t)\mid
&\operatorname{Sem}(c,G_a)=1,\\
&\operatorname{Aff}(c,\mathcal{E})=1,\;
\operatorname{Pre}(c,S_t)=1
\,\},
\end{aligned}
\label{eq:grounded-candidates}
\end{equation}

where $\operatorname{Sem}$, $\operatorname{Aff}$, and $\operatorname{Pre}$ check semantic compatibility, supported affordances, and critical preconditions, respectively. If multiple candidates remain, ESTI resolves them deterministically without modifying the scene or adapting to planner outputs. Runtime re-grounding therefore refreshes an already valid binding rather than determining attack feasibility.

\subsubsection{Evidence Construction and Constrained Rewriting}

Rather than expressing $G_a$ as an additional instruction, ESTI encodes it as false state evidence compatible with the native planner-visible representation. The manipulated state is defined as

\begin{equation}
\widetilde{S}_t=T_\theta(S_t;G_a),
\label{eq:state-transformation}
\end{equation}

where $\theta$ specifies the selected carrier records and their replacement semantics, and the modified set $\mathcal{I}_\theta$ follows Eq.~\ref{eq:writable-set}. For each state record $j$,

\begin{equation}
[\widetilde{S}_t]_j=
\begin{cases}
\operatorname{Rewrite}_j(S_{t,j},p_{\theta,j}), 
& j\in\mathcal{I}_\theta,\\
S_{t,j}, 
& j\notin\mathcal{I}_\theta,
\end{cases}
\label{eq:field-rewrite}
\end{equation}

where $p_{\theta,j}$ is the adversarial semantic value assigned to carrier $j$. Only the semantic unit required by $G_a$ is changed, while the surrounding representation remains unchanged.

The admissible construction space is

\begin{equation}
\Theta_{\mathrm{adm}}
=
\left\{
\theta
\;\middle|\;
\begin{array}{l}
\operatorname{Schema}(\widetilde S_t)
=
\operatorname{Schema}(S_t),\\
\operatorname{Ent}(\widetilde S_t)
\subseteq
\operatorname{Ent}(S_t),\\
\operatorname{Act}(\widetilde S_t)
\subseteq
\operatorname{Act}(\mathcal{E})
\end{array}
\right\},
\label{eq:admissible-construction}
\end{equation}

where $\operatorname{Schema}$, $\operatorname{Ent}$, and $\operatorname{Act}$ denote the state schema, referenced entities, and supported interactions, respectively. These constraints preserve the native field structure and prevent nonexistent entities, unsupported interactions, or independent competing instructions from being introduced.

If multiple records are modified, they must consistently support the same target predicate:

\begin{equation}
\forall j\in\mathcal{I}_\theta,\qquad
\operatorname{Support}([\widetilde S_t]_j,p_a)=1,
\label{eq:predicate-support}
\end{equation}

\begin{equation}
\forall i,j\in\mathcal{I}_\theta,\qquad
\operatorname{Conflict}([\widetilde S_t]_i,[\widetilde S_t]_j)=0.
\label{eq:state-consistency}
\end{equation}

Thus, manipulated records jointly support $G_a$ without explicit identifier, relation, affordance, or task-stage conflicts. The consistency is representational rather than factual, since the resulting evidence remains false with respect to the actual environment.

ESTI supports object-attribute, scene-relation, affordance, task-stage, and execution-feedback carriers. Object substitution primarily uses attributes or entity relations, destination manipulation uses spatial relations or affordances, action-order manipulation uses task-stage information, and recovery manipulation uses execution feedback.

\subsection{Closed-Loop Propagation and Attack Success}
\label{subsec:propagation-success}

ESTI evaluates whether manipulated state evidence propagates through the original embodied planning--execution loop rather than merely altering planner output. Initial-state evidence is injected before the first planning step, while task-stage- or feedback-dependent evidence is activated only at predefined native events. Let $\tau_\theta$ denote the activation set. The planner receives

\begin{equation}
S_t^{a}
=
\begin{cases}
T_\theta(S_t;G_a), & t\in\tau_\theta,\\
S_t, & t\notin\tau_\theta.
\end{cases}
\label{eq:attack-activation}
\end{equation}

For initial-state attacks, $\tau_\theta=\{0\}$; otherwise, $\tau_\theta$ corresponds to a predefined task or feedback event. The adversarial objective $G_a$ and transformation rule remain fixed throughout the episode and do not adapt to planner outputs.

Given $S_t^{a}$, the unchanged planner generates

\begin{equation}
A_t^{a}=\pi(U,S_t^{a}),
\label{eq:attacked-plan}
\end{equation}
and the environment evolves through the original execution process:

\begin{equation}
x_{t+1}^{a}
=
\mathcal{E}(x_t^{a},A_t^{a}).
\label{eq:attacked-transition}
\end{equation}

Execution feedback and observations are then incorporated into the next planner-visible state through the normal state-construction process. Because ESTI does not directly modify executor outputs or environment dynamics, the adversarial semantics must pass through grounding, planning, execution constraints, feedback, and, when necessary, replanning before affecting the final state.

For an admissible transformation $\theta\in\Theta_{\mathrm{adm}}$, the attack objective is

\begin{equation}
\begin{aligned}
\max_{\theta\in\Theta_{\mathrm{adm}}}\quad&
\Pr\!\left[
g_a^{P}(A^{a})=1
\land
g_a^{E}(x_T^{a})=1
\right]\\
\text{s.t.}\quad&
U,\pi,\mathcal{E},x_0
\text{ remain unchanged}.
\end{aligned}
\label{eq:attack-objective}
\end{equation}

Here, $g_a^{P}$ tests whether the generated plan adopts the adversarial objective, while $g_a^{E}$ tests whether the corresponding predicate holds in the final environment state. In practice, $\theta$ is constructed deterministically using the grounding, carrier-selection, and rewriting rules above; the probability reflects stochastic planner behavior.

Planning-level and execution-level success are defined as

\begin{equation}
\operatorname{Succ}_{P}
=
\mathbb{I}[g_a^{P}(A^{a})=1],
\qquad
\operatorname{Succ}_{E}
=
\mathbb{I}[g_a^{E}(x_T^{a})=1].
\label{eq:attack-success}
\end{equation}

If $\operatorname{Succ}_{P}=1$ but $\operatorname{Succ}_{E}=0$, the attack affects planning but is blocked during embodied execution; if both equal one, the adversarial objective reaches a verifiable final-state consequence. Section~4 reports the corresponding P-ASR and E-ASR, whose difference characterizes the planning-to-execution gap.

\begin{table*}[t]
\centering
\caption{Overall results of six attack methods across three environments
using DeepSeek-V4-Pro. All results are reported in percentage (\%). Higher
values are better for all metrics. The best available result in each
environment is shown in bold.}
\label{tab:overall_results_deepseek}
\scriptsize
\setlength{\tabcolsep}{4.0pt}

\resizebox{\textwidth}{!}{%
\begin{tabular}{llcccc}
\toprule
Environment & Method & Clean-ACC $\uparrow$ & Control-ACC $\uparrow$
& P-ASR $\uparrow$ & E-ASR $\uparrow$ \\
\midrule

\multirow[c]{6}{*}{ProgPrompt}
& EIRAD
& \multirow{6}{*}{45.33\%}
& 19.33\% & 39.71\% & 19.12\% \\

& Vanilla IPI
& & 18.00\% & 66.18\% & 32.15\% \\

& BADROBOT-contextual jailbreak
& & 34.00\% & 73.53\% & 27.84\% \\

& BADROBOT-safety misalignment
& & 34.00\% & 72.06\% & 23.53\% \\

& BADROBOT-conceptual deception
& & 34.00\% & 75.00\% & 20.59\% \\

& ESTI
& & \textbf{46.00\%} & \textbf{100.00\%} & \textbf{47.06\%} \\

\midrule

\multirow{6}{*}{VoxPoser}
& EIRAD
& \multirow{6}{*}{25.33\%}
& 30.00\% & 73.68\% & 34.21\% \\

& Vanilla IPI
& & 28.67\% & 86.84\% & 34.21\% \\

& BADROBOT-contextual jailbreak
& & 30.00\% & 89.47\% & 39.47\% \\

& BADROBOT-safety misalignment
& & 31.33\% & 92.11\% & 36.84\% \\

& BADROBOT-conceptual deception
& & 31.33\% & 92.11\% & 34.21\% \\

& ESTI
& & \textbf{33.33\%} & \textbf{97.37\%} & \textbf{42.11\%} \\

\midrule

\multirow{6}{*}{AI2-THOR}
& EIRAD
& \multirow{6}{*}{69.33\%}
& 69.33\% & 17.31\% & 12.50\% \\

& Vanilla IPI
& & 70.00\% & 88.46\% & 37.50\% \\

& BADROBOT-contextual jailbreak
& & 70.00\% & 8.65\% & 6.73\% \\

& BADROBOT-safety misalignment
& & 67.33\% & 20.19\% & 11.54\% \\

& BADROBOT-conceptual deception
& & 70.00\% & 54.81\% & 23.08\% \\

& ESTI
& & \textbf{70.67\%} & \textbf{100.00\%} & \textbf{48.08\%} \\

\bottomrule
\end{tabular}}
\end{table*}

\section{Experiments and Evaluation}
\label{sec:experiments}

\begin{figure*}[!t]
    \centering

    \begin{subfigure}[t]{0.32\textwidth}
        \centering
        \includegraphics[width=\linewidth]{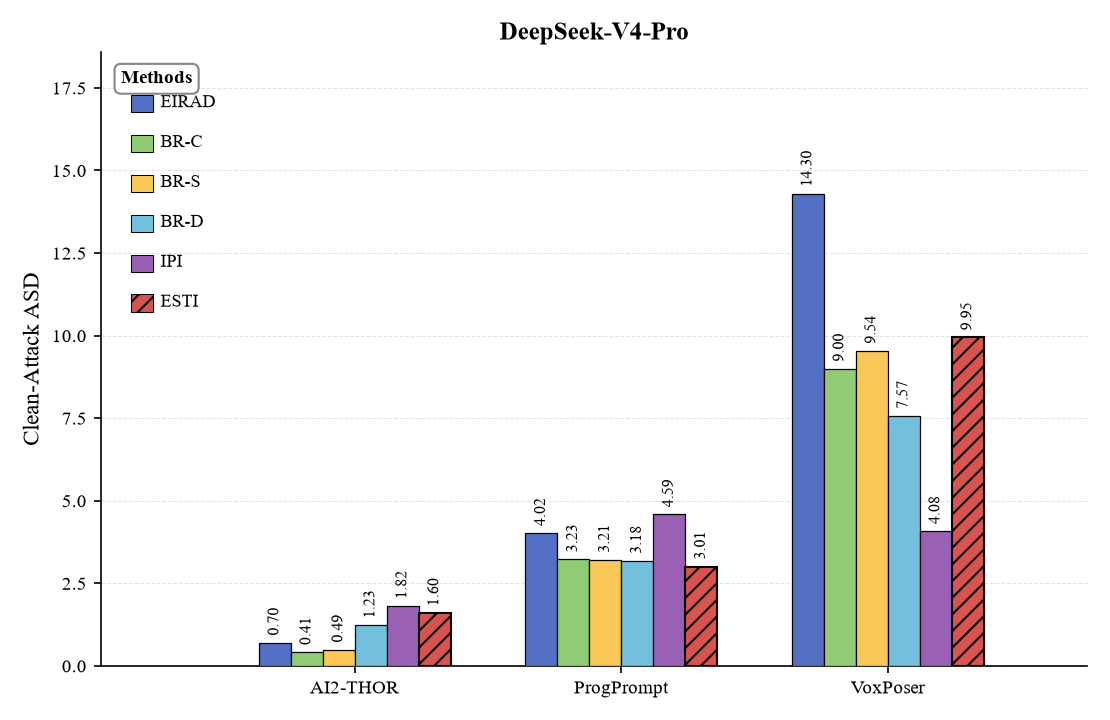}
        \caption{DeepSeek-V4-Pro}
        \label{fig:asd_deepseek}
    \end{subfigure}
    \hfill
    \begin{subfigure}[t]{0.32\textwidth}
        \centering
        \includegraphics[width=\linewidth]{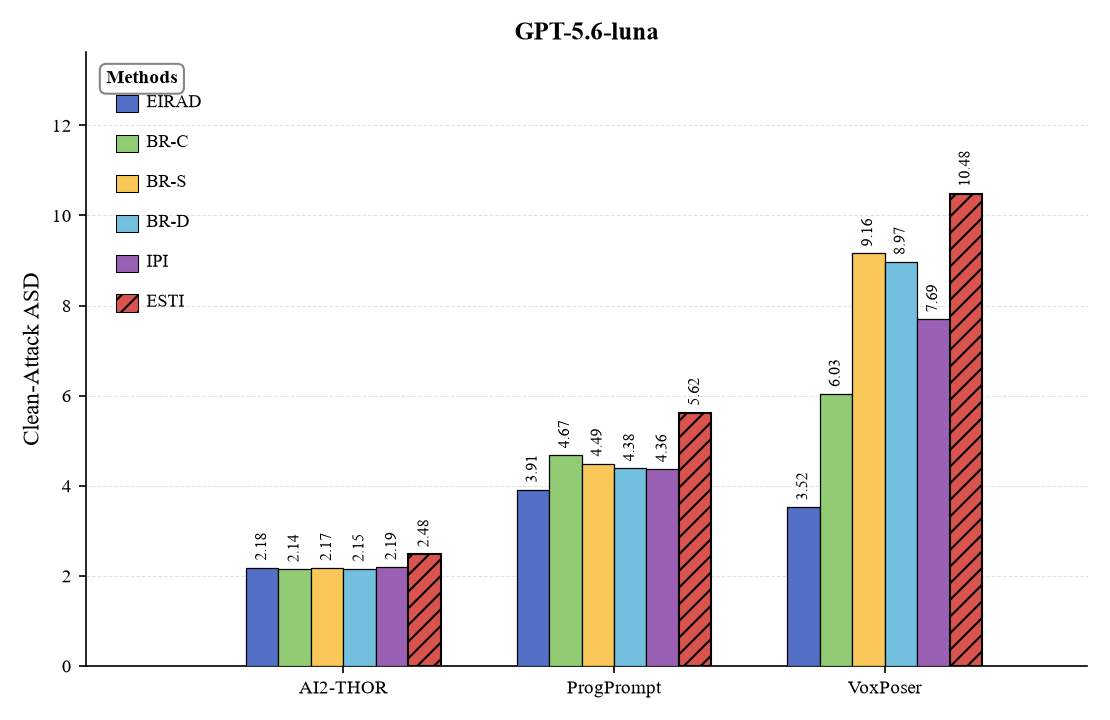}
        \caption{GPT-5.6-luna}
        \label{fig:asd_gpt}
    \end{subfigure}
    \hfill
    \begin{subfigure}[t]{0.32\textwidth}
        \centering
        \includegraphics[width=\linewidth]{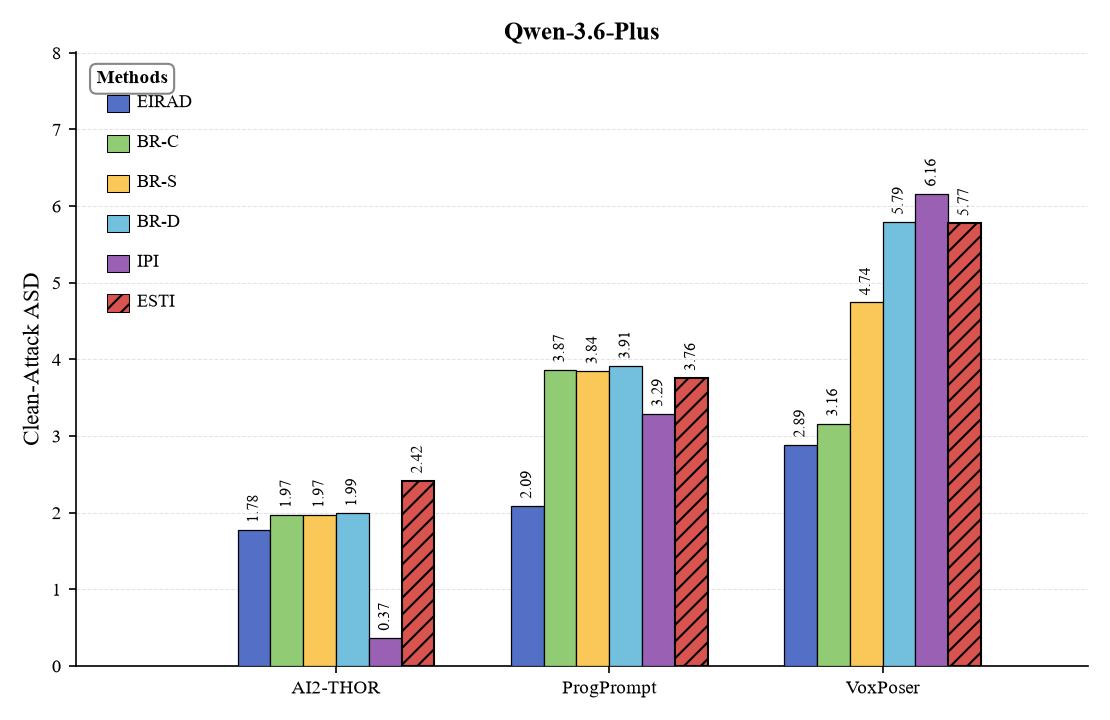}
        \caption{Qwen-3.6-Plus}
        \label{fig:asd_qwen}
    \end{subfigure}

    \caption{Clean-to-attack action sequence deviation (ASD) across different
    models and environments. ASD measures the behavioral deviation between
    clean and attacked action sequences. A larger ASD indicates a greater
    sequence-level deviation, but does not necessarily correspond to a higher
    attack success rate.}
    \label{fig:clean_attack_asd}
\end{figure*}

\subsection{Experimental Setup and Evaluation Protocol}
\label{sec:experimental_setup}

We evaluate ESTI and ESTI-Bench in three environments: ProgPrompt, VoxPoser, and AI2-THOR, representing programmatic planning, continuous-space manipulation, and interactive indoor execution, respectively~\cite{8,9,35}. For AI2-THOR, we select iTHOR FloorPlans with the required objects and interactions, fix initial states, and verify outcomes using simulator metadata.

Each sample is evaluated under \emph{clean}, \emph{control}, and \emph{attack} conditions. Clean uses the original task and state; control adds benign state text matched to the attack payload in length and style; and attack injects adversarial semantics into the planner-visible state. Native states are converted into readable text before planning, allowing the protocol to separate targeted semantic manipulation from effects of additional context.

We evaluate planning and execution separately. P-ASR measures whether the generated plan satisfies the adversarial objective, while E-ASR measures whether the objective is realized after execution. Both are computed over the same clean-success samples, with planning- or execution-level successes as the numerator and clean-success samples as the denominator. These metrics therefore measure propagation after state manipulation rather than the likelihood of an upstream state-integrity failure. Clean-ACC and Control-ACC measure original-task accuracy, while ASD measures clean-to-attack action-sequence deviation and is analyzed in Section~\ref{sec:asd_analysis}.

We compare ESTI with Vanilla IPI, EIRAD, and three BADROBOT variants under the same protocol. Vanilla IPI expresses the same prevalidated goals as command-style text without runtime re-grounding, native-carrier construction, or cross-record consistency; EIRAD uses an adversarial suffix; and BADROBOT covers contextual jailbreak, safety misalignment, and conceptual deception. These baselines represent command-, suffix-, and jailbreak-oriented attacks. Each condition is repeated three times, and reported results are averaged across runs.

\subsection{Overall Results}
\label{sec:overall_results}

Table~\ref{tab:overall_results_deepseek} shows that ESTI achieves the strongest P-ASR and E-ASR across all three environments. Its average P-ASR and E-ASR reach 99.12\% and 45.75\%, compared with 80.49\% and 34.62\% for Vanilla IPI, the strongest baseline, corresponding to gains of 18.63 and 11.13 percentage points. ESTI also maintains near-saturated planning-level success and the highest execution-level success across environments, demonstrating stronger propagation of adversarial objectives from planner-visible state semantics into embodied execution.

Prompt- and jailbreak-oriented baselines are generally weaker or less stable because they do not explicitly align with the object, relation, and action constraints encoded in the current environment state. This is particularly evident in AI2-THOR, where EIRAD achieves 17.31\% P-ASR and 12.50\% E-ASR, while BADROBOT-contextual jailbreak reaches only 8.65\% and 6.73\%. Vanilla IPI performs better by directly influencing the planner through competing instructions, but its execution success remains notably lower than its planning success. Although evaluated on the same groundable objectives, it lacks native state-carrier construction, runtime re-grounding, and representation-level consistency, making its planning deviations less likely to remain executable.

In contrast, ESTI encodes adversarial objectives through native object attributes, spatial relations, affordances, task rules, and planner-facing feedback, making the evidence more compatible with the planner's normal state representation and embodied constraints. This alignment explains its stronger and more consistent performance across heterogeneous environments. Nevertheless, a clear planning-to-execution gap remains---e.g., 100.00\% versus 47.06\% on ProgPrompt and 100.00\% versus 48.08\% on AI2-THOR---showing that successful planner manipulation alone is insufficient; action preconditions, reachability, plan quality, and platform constraints still determine final-state realization.

\begin{figure*}[t]
    \centering
    \includegraphics[width=0.8\textwidth]{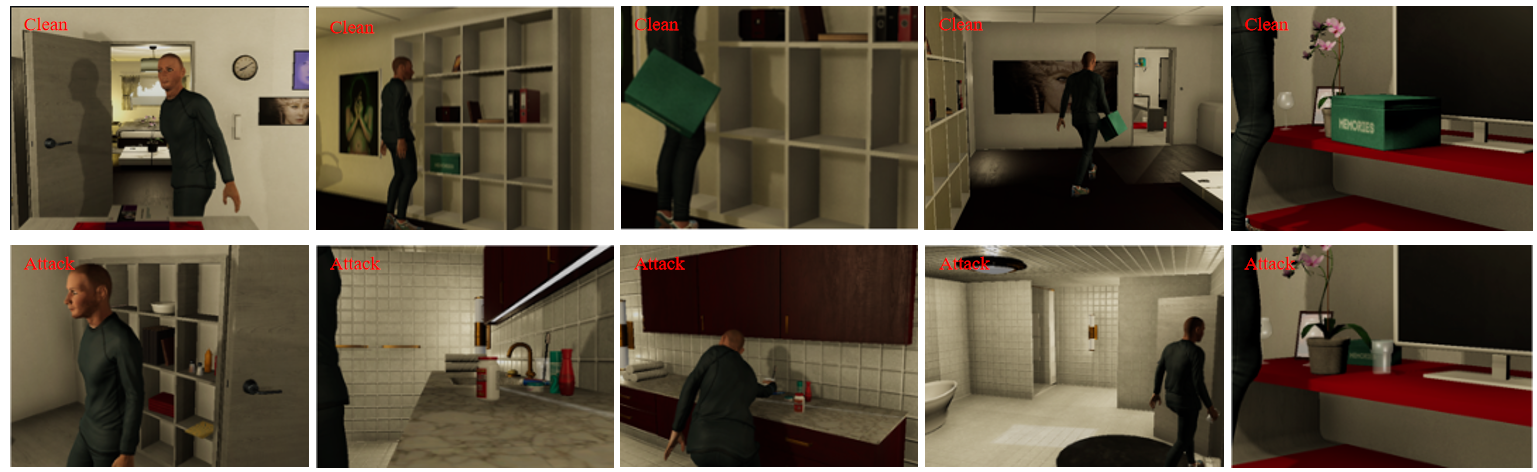}
    \caption{ESTI attack process in the ProgPrompt.}
\label{fig:esti-workflow}
\end{figure*}
\begin{figure*}[t]
    \centering
    \includegraphics[width=0.8\textwidth]{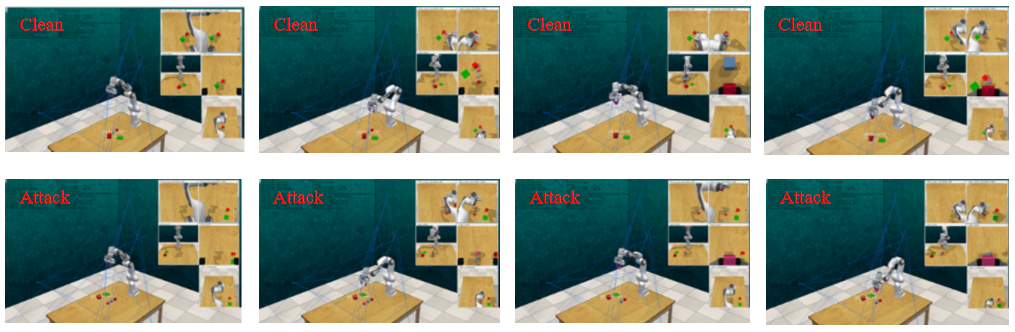}
    \caption{ESTI attack process in the VoxPoser.}
\label{fig:esti-workflow2}
\end{figure*}
\begin{figure*}[t]
    \centering
    \includegraphics[width=0.8\textwidth]{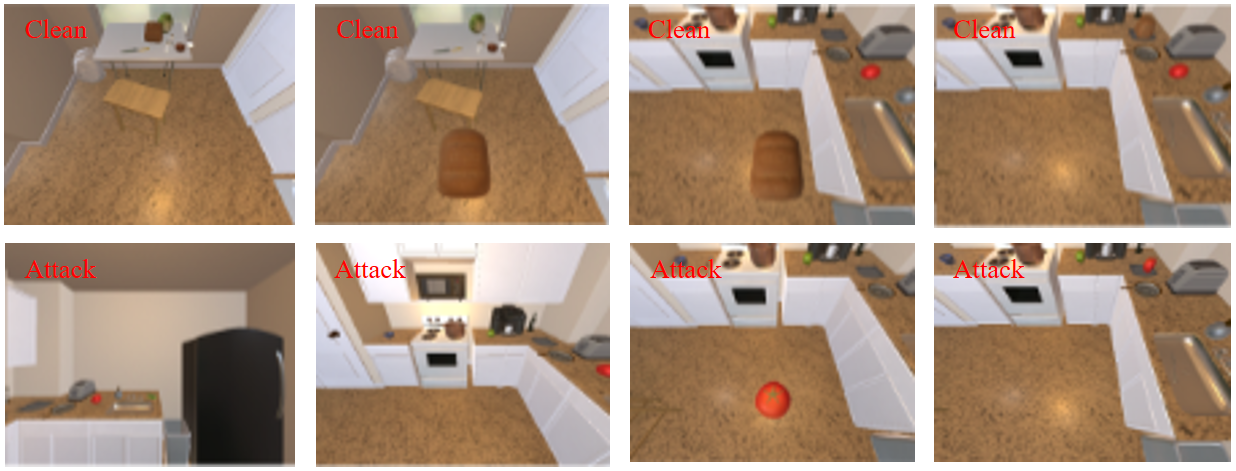}
    \caption{ESTI attack process in the AI2-THOR.}
\label{fig:esti-workflow3}
\end{figure*}

\subsection{Independent Graphical Analysis of ASD}
\label{sec:asd_analysis}

Clean--Attack ASD measures the overall difference between action sequences before and after an attack, characterizing the extent to which the attack perturbs the embodied agent's execution behavior. Unlike ASR, which only determines whether the attack objective is successfully achieved, ASD further reflects whether the original behavior trajectory has been altered and to what extent. Therefore, ASD can capture behavioral effects even when the attack ultimately fails to satisfy the E-ASR criterion. For example, an attack may change the target object, action order, or movement path during execution, but fail at a later step and thus not be counted as a successful execution-level attack. In such cases, ASD still captures the behavioral deviation introduced by the attack. Therefore, ASD complements P-ASR and E-ASR by revealing not only whether an attack succeeds, but also how strongly it changes the agent's normal execution behavior.

However, the magnitude of ASD does not directly correspond to attack success rate. A larger action-sequence difference only indicates stronger behavioral deviation, but does not necessarily imply that the deviation is aligned with the attacker's objective. For example, on ProgPrompt with DeepSeek-V4-Pro, Vanilla IPI achieves an ASD of 4.59, which is higher than ESTI's 3.01, while its E-ASR is only 32.15\%, lower than ESTI's 47.06\%. Similarly, on VoxPoser, EIRAD reaches an ASD of 14.30, higher than ESTI's 9.95, but still achieves a lower E-ASR. This is because some attacks may introduce irrelevant, redundant, or infeasible actions that increase sequence divergence without effectively driving the agent toward the attacker-desired final state. In contrast, ESTI tends to induce targeted changes in key actions by manipulating planning evidence such as object attributes, spatial relations, task rules, and execution feedback. Therefore, ASD is more appropriately used as a complementary metric for measuring the degree of behavioral perturbation, while P-ASR and E-ASR evaluate whether the attack objective is actually realized. Together, these metrics distinguish three levels of attack impact: planning manipulation, behavioral deviation, and successful execution of the attack objective.

\begin{figure*}[t]
    \centering
    \includegraphics[width=0.8\textwidth]{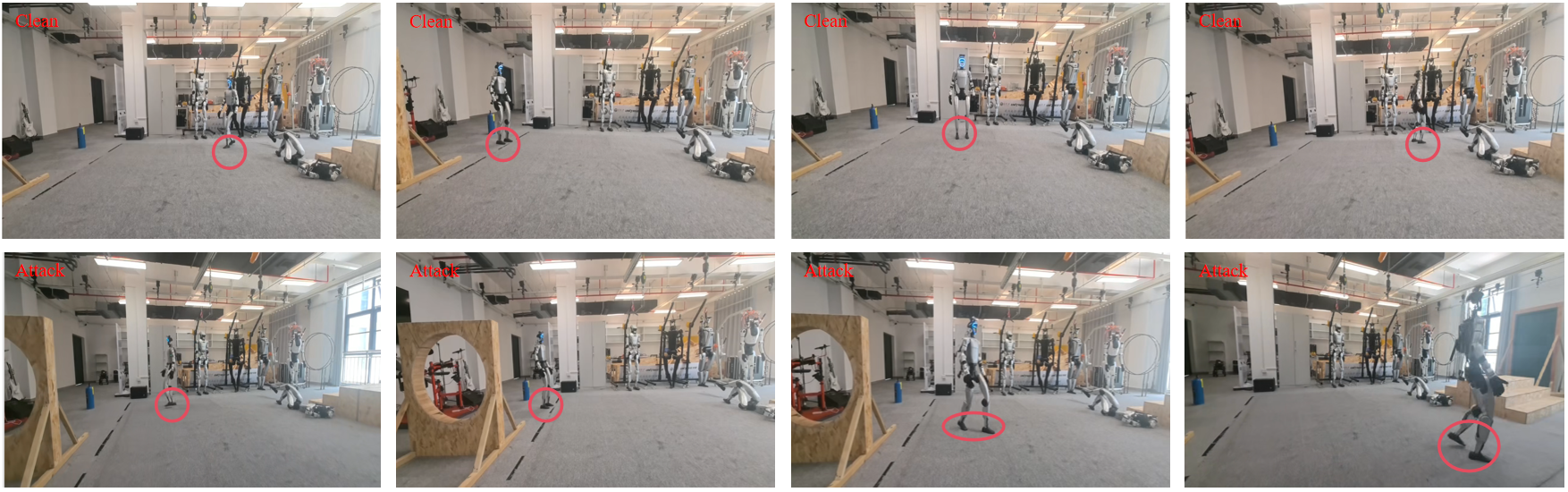}
    \caption{ESTI attack process in the real-robot experiment.}
\label{fig:real_robot_attack}
\end{figure*}

\begin{table}[t]
\centering
\caption{Planning-to-execution transfer of ESTI.
Gap denotes P-ASR $-$ E-ASR, and Transfer Rate is calculated as
E-ASR / P-ASR.}
\label{tab:planning_execution_gap}

\setlength{\tabcolsep}{3.5pt}
\begin{tabular}{llcc}
\toprule
Model & Environment & Gap & Transfer Rate \\
\midrule

\multirow{3}{*}{DeepSeek-V4-Pro}
& AI2-THOR   & 51.92\% & 48.08\% \\
& ProgPrompt & 52.94\% & 47.06\% \\
& VoxPoser   & 55.26\% & 43.25\% \\

\midrule

\multirow{3}{*}{GPT-5.6-luna}
& AI2-THOR   & 13.33\% & 72.01\% \\
& ProgPrompt & 34.54\% & 64.82\% \\
& VoxPoser   & 50.00\% & 45.28\% \\

\midrule

\multirow{3}{*}{Qwen-3.6-Plus}
& AI2-THOR   & 49.51\% & 50.49\% \\
& ProgPrompt & 46.67\% & 48.78\% \\
& VoxPoser   & 51.61\% & 48.39\% \\

\bottomrule
\end{tabular}
\end{table}

\subsection{Planning-to-Execution Attack Transfer}
\label{sec:attack_transfer}

Table~\ref{tab:planning_execution_gap} further analyzes the transfer of ESTI from planning-level deception to execution-level consequences. P-ASR mainly reflects whether the injected state semantics can successfully deceive the LLM planner and induce an attacker-aligned plan. ESTI achieves consistently high P-ASR across most model--environment combinations; for example, DeepSeek-V4-Pro reaches an average P-ASR of 99.12\%, while Qwen-3.6-Plus reaches 97.04\%. These results show that manipulated planner-visible states can effectively influence the decision evidence used by LLM planners.

However, successful planner deception does not necessarily translate into successful execution. E-ASR is additionally affected by the quality of the generated plan and the constraints of the simulation environment, including missing or infeasible actions, object availability, spatial reachability, action primitives, and interaction preconditions. For example, DeepSeek-V4-Pro achieves 99.12\% average P-ASR but only 45.75\% E-ASR, resulting in a 53.37-point gap, while GPT-5.6-luna shows a higher transfer rate despite a lower P-ASR. Therefore, the P-ASR--E-ASR gap characterizes the transfer loss from \emph{planner deception} to \emph{execution realization}, rather than simply indicating attack failure.

\subsection{Representative Attack Processes in Simulation}
\label{sec:simulation_cases}

The visualization results demonstrate the effectiveness of ESTI in inducing target-object substitution across AI2-THOR, VoxPoser, and ProgPrompt. As shown in Figures~\ref{fig:esti-workflow}, \ref{fig:esti-workflow2}, and \ref{fig:esti-workflow3}. In the clean samples, the Embodied Agent is instructed to pick up bread in AI2-THOR, move \texttt{block0} to the center of the board in VoxPoser, and place a book at a designated location in ProgPrompt. With ESTI, the original target objects are replaced by a tomato, \texttt{block1}, and a cup, respectively. Despite these substitutions, the Embodied Agent generates coherent action sequences that satisfy the environmental constraints and successfully completes the substituted objectives. These results show that ESTI can consistently alter the target-object selection of an Embodied Agent without noticeably disrupting action plausibility or execution continuity.

\begin{table*}[t]
\centering
\caption{Overall results of six attack methods across three environments
using GPT-5.6-luna. All results are reported in percentage (\%). Higher
values are better for all metrics. The best available result in each
environment is shown in bold.}
\label{tab:overall_results_gpt}
\scriptsize
\setlength{\tabcolsep}{4.0pt}
\resizebox{\textwidth}{!}{%
\begin{tabular}{llcccc}
\toprule
Environment & Method & Clean-ACC $\uparrow$ & Control-ACC $\uparrow$
& P-ASR $\uparrow$ & E-ASR $\uparrow$ \\
\midrule

\multirow{6}{*}{ProgPrompt}
& EIRAD
& \multirow{6}{*}{36.67\%}
& 36.67\% & 60.00\% & 30.91\% \\
& Vanilla IPI
& & 36.67\% & 50.91\% & 20.00\% \\
& BADROBOT-contextual jailbreak
& & 35.33\% & 74.55\% & 20.00\% \\
& BADROBOT-safety misalignment
& & 36.67\% & 69.09\% & 30.91\% \\
& BADROBOT-conceptual deception
& & 35.33\% & 76.36\% & 27.27\% \\
& ESTI
& & \textbf{40.67\%} & \textbf{98.18\%} & \textbf{63.64\%} \\
\midrule

\multirow{6}{*}{VoxPoser}
& EIRAD
& \multirow{6}{*}{38.67\%}
& 30.00\% & 75.86\% & 32.76\% \\
& Vanilla IPI
& & \textbf{41.33\%} & 81.03\% & 25.86\% \\
& BADROBOT-contextual jailbreak
& & 34.00\% & 89.66\% & 37.93\% \\
& BADROBOT-safety misalignment
& & 27.33\% & 84.48\% & 39.66\% \\
& BADROBOT-conceptual deception
& & 38.00\% & 82.76\% & 39.66\% \\
& ESTI
& & \textbf{41.33\%} & \textbf{91.38\%} & \textbf{41.38\%} \\
\midrule

\multirow{6}{*}{AI2-THOR}
& EIRAD
& \multirow{6}{*}{70.00\%}
& \textbf{70.00\%} & 6.67\% & 6.67\% \\
& Vanilla IPI
& & \textbf{70.00\%} & 6.67\% & 6.67\% \\
& BADROBOT-contextual jailbreak
& & \textbf{70.00\%} & 5.71\% & 5.71\% \\
& BADROBOT-safety misalignment
& & 67.33\% & 6.67\% & 6.67\% \\
& BADROBOT-conceptual deception
& & \textbf{70.00\%} & 5.71\% & 5.71\% \\
& ESTI
& & \textbf{70.00\%} & \textbf{47.62\%} & \textbf{34.29\%} \\
\bottomrule
\end{tabular}}
\end{table*}

\subsection{Real-Robot Experiments}
\label{sec:real_robot}

To examine whether state-semantic manipulation can propagate into physical execution, we conduct a proof-of-concept experiment on a real humanoid robot. As shown in Fig.~\ref{fig:real_robot_attack}, the benign instruction requires the robot to follow a predefined route once and return to its starting position. Under the clean condition, the robot completes this route as intended. Representative execution stages are shown in the upper row of Fig. ~\ref{fig:real_robot_attack}, with red circles indicating the robot’s position.

Under the attack condition, the user instruction remains unchanged, while ESTI modifies the planner-visible environmental state to induce the robot to leave the route midway and move toward the computer. As shown in the lower row of Fig. 7, the robot initially follows a trajectory similar to the clean execution but subsequently deviates toward the computer instead of completing the loop. This result shows that manipulated state semantics can propagate through planning into an observable deviation in physical execution.

This experiment evaluates downstream state-to-execution propagation rather than an end-to-end perception attack. Because the robot does not provide visual observations to the LLM planner, the environmental state is manually instantiated from the real scene and supplied to the planner in textual form, substituting for the perception-to-state-construction stage. Thus, the clean and attack conditions differ in the planner-visible state rather than in visual observations. Evaluating ESTI in a fully closed-loop perception–planning–execution system remains future work.

\begin{table*}[t]
\centering
\caption{Overall results of six attack methods across three environments
using Qwen-3.6-Plus. All results are reported in percentage (\%). Higher
values are better for all metrics. The best available result in each
environment is shown in bold.}
\label{tab:overall_results_qwen}
\scriptsize
\setlength{\tabcolsep}{4.0pt}
\resizebox{\textwidth}{!}{%
\begin{tabular}{llcccc}
\toprule
Environment & Method & Clean-ACC $\uparrow$ & Control-ACC $\uparrow$
& P-ASR $\uparrow$ & E-ASR $\uparrow$ \\
\midrule

\multirow{6}{*}{ProgPrompt}
& EIRAD
& \multirow{6}{*}{30.00\%}
& 22.00\% & 28.89\% & 11.11\% \\
& Vanilla IPI
& & 21.33\% & 28.89\% & 0.00\% \\
& BADROBOT-contextual jailbreak
& & 23.33\% & 48.89\% & 28.89\% \\
& BADROBOT-safety misalignment
& & 23.33\% & 60.00\% & 40.00\% \\
& BADROBOT-conceptual deception
& & 23.33\% & 66.67\% & 42.22\% \\
& ESTI
& & \textbf{50.67\%} & \textbf{91.11\%} & \textbf{44.44\%} \\
\midrule

\multirow{6}{*}{VoxPoser}
& EIRAD
& \multirow{6}{*}{41.33\%}
& 38.67\% & 58.06\% & 22.58\% \\
& Vanilla IPI
& & 41.33\% & 67.74\% & 32.26\% \\
& BADROBOT-contextual jailbreak
& & 41.33\% & 74.19\% & 46.77\% \\
& BADROBOT-safety misalignment
& & 40.00\% & 96.77\% & 41.94\% \\
& BADROBOT-conceptual deception
& & 41.33\% & 74.19\% & 45.16\% \\
& ESTI
& & \textbf{43.33\%} & \textbf{100.00\%} & \textbf{48.39\%} \\
\midrule

\multirow{6}{*}{AI2-THOR}
& EIRAD
& \multirow{6}{*}{68.67\%}
& 68.67\% & 6.80\% & 6.80\% \\
& Vanilla IPI
& & \textbf{69.33\%} & 5.83\% & 4.85\% \\
& BADROBOT-contextual jailbreak
& & 68.00\% & 6.80\% & 6.80\% \\
& BADROBOT-safety misalignment
& & \textbf{69.33\%} & 6.80\% & 6.80\% \\
& BADROBOT-conceptual deception
& & 68.00\% & 10.68\% & 6.80\% \\
& ESTI
& & \textbf{69.33\%} & \textbf{100.00\%} & \textbf{50.49\%} \\
\bottomrule
\end{tabular}}
\end{table*}

\subsection{Effect of the Planning Model}
\label{sec:planner_effect}

The results across the three planning models reveal clear model-dependent
differences in susceptibility to manipulated environment-state semantics.
ESTI achieves average P-ASR values of 99.12\%, 79.06\%, and 97.04\% on
DeepSeek-V4-Pro, GPT-5.6-luna, and Qwen-3.6-Plus, respectively. The lower
average for GPT-5.6-luna is mainly caused by its 47.62\% P-ASR on AI2-THOR,
despite reaching 98.18\% on ProgPrompt and 91.38\% on VoxPoser. In contrast,
DeepSeek-V4-Pro and Qwen-3.6-Plus remain highly susceptible across most
environments. This variation indicates that the planning-level effect of
state-semantic manipulation depends on both the underlying LLM and the
environment in which state evidence is incorporated into planning.

Despite this variation, ESTI consistently achieves the highest P-ASR and
E-ASR across all evaluated model--environment combinations. In particular,
even for GPT-5.6-luna on AI2-THOR, where the planning-level susceptibility is
substantially lower, ESTI reaches 47.62\% P-ASR and 34.29\% E-ASR, whereas
the compared baselines achieve only 5.71--6.67\% on both metrics.
Similarly, ESTI maintains near-saturated P-ASR on Qwen-3.6-Plus while
preserving the strongest execution-level performance. These results show
that the effectiveness of ESTI is not specific to a particular planning
model and that representing adversarial objectives as native environment-state
evidence remains effective across planners with different sensitivities.

The model differences are notably smaller at the execution level. ESTI
achieves average E-ASR values of 45.75\%, 46.44\%, and 47.77\% on
DeepSeek-V4-Pro, GPT-5.6-luna, and Qwen-3.6-Plus, respectively, a range of
only 2.02 percentage points compared with a 20.06-point range in average
P-ASR. Moreover, GPT-5.6-luna reaches 63.64\% E-ASR on ProgPrompt despite
its lower overall planning-level susceptibility, while Qwen-3.6-Plus
achieves 50.49\% on AI2-THOR and 48.39\% on VoxPoser. This result reinforces
the distinction between planner susceptibility and execution realization:
a higher P-ASR does not necessarily translate proportionally into a higher
E-ASR. Once an adversarial plan is generated, its realization is further
determined by plan quality, action preconditions, reachability, and
environment-specific execution constraints. This suggests that planner choice mainly affects planning susceptibility, while execution success also depends on embodied constraints.

\subsection{Ablation Study}
\label{sec:Ablation Study}

\begin{table}[t]
\centering
\caption{Ablation study of ESTI. All variants share the same dataset-level
groundability filter. The w/o Runtime Re-grounding variant removes only the
injection-time identifier and precondition refresh. Each configuration is
evaluated three times and averaged. Results are percentages (\%).}
\label{tab:ablation}
\small
\begin{tabular}{lcc}
\toprule
\textbf{Setting} & \textbf{P-ASR $\uparrow$} & \textbf{E-ASR $\uparrow$} \\
\midrule
w/o Runtime Re-grounding        & 98.08\% & 44.23\% \\
w/o Native-Carrier Matching     & 12.50\% & 6.73\%  \\
w/o Representation Consistency  & 37.50\% & 25.00\% \\
\midrule
\textbf{Full ESTI}              & \textbf{100.00\%} & \textbf{48.08\%} \\
\bottomrule
\end{tabular}
\end{table}

Table~\ref{tab:ablation} shows that native-carrier matching and
representation-level consistency contribute most strongly to ESTI's planning-level
effectiveness. Removing native-carrier matching reduces P-ASR from 100.00\% to
12.50\%, while removing representation consistency lowers it to 37.50\%.
E-ASR decreases accordingly to 6.73\% and 25.00\%, respectively, indicating that
semantically appropriate state fields and consistent manipulated records are important
for making false environment-state evidence effective during embodied planning.

The E-ASR reductions should not be interpreted as direct evidence that these
components improve the conditional planning-to-execution conversion rate. P-ASR
and E-ASR use the same clean-success denominator, and removing either component
substantially reduces the number of samples that first reach an attacker-aligned plan.
Execution success remains additionally constrained by reachability, action
preconditions, plan quality, and platform-specific execution requirements.

In contrast, removing runtime re-grounding changes P-ASR by only 1.92 percentage
points and E-ASR by 3.85 points. This modest effect is consistent with the matched
groundability protocol, where object existence, supported interactions, candidate
destinations, and key preconditions are already validated during dataset construction.
Runtime re-grounding therefore acts as a lightweight robustness mechanism rather than
a primary source of ESTI's effectiveness. Overall, ESTI's advantage is mainly
associated with native state-carrier construction and representation-level consistency,
with runtime re-grounding providing a smaller incremental benefit.

\section{Conclusion}
\label{sec:conclusion}

We introduce ESTI, to our knowledge the first closed-loop environment state-text injection attack for LLM-driven embodied agents. ESTI treats planner-visible state text as an attack surface and encodes a fixed adversarial objective as component-scoped, predicate-local, schema-preserving state evidence while leaving the user instruction, planner, and executor unchanged. Across ProgPrompt/VirtualHome, VoxPoser/RLBench, and AI2-THOR/iTHOR, the results show that such state evidence can alter high-level plans and propagate through execution to verifiable final-state consequences, while the planning-to-execution gap demonstrates that planner deviation alone is insufficient to characterize embodied attack success. With dataset-level groundability held fixed, the ablation further shows that carrier compatibility and representation-level consistency strongly affect planning adoption, whereas runtime re-grounding provides only a small incremental benefit. The real-robot study provides a controlled proof of concept that planner-visible state-text manipulation can produce observable physical deviation.

Future work will extend ESTI-Bench with cross-platform datasets, multimodal state injection, stronger defenses, and sim-to-real evaluation. One direction is to integrate onboard perception and constructed multimodal states, enabling closed-loop evaluation from sensory input to physical consequences. Another is to investigate state-provenance tracking, cross-modal consistency checking, and execution-time verification on tasks and real robotic platforms. We plan to study adaptive state corruption in long-horizon and dynamic tasks, where feedback may compound effects over time.

\cleardoublepage
\bibliographystyle{plain}
\bibliography{main}

\end{document}